\documentclass{article}

\usepackage[preprint,nonatbib]{neurips_2020}

\usepackage[utf8]{inputenc} 
\usepackage[T1]{fontenc}    
\usepackage{hyperref}       
\usepackage{url}            
\usepackage{booktabs}       
\usepackage{amsfonts}       
\usepackage{nicefrac}       
\usepackage{microtype}      
\usepackage{amssymb}
\usepackage{amssymb}
\usepackage{amsmath}
\usepackage[pdftex]{graphicx}
\usepackage{caption} 
\usepackage{subcaption}
\usepackage{placeins}
\usepackage{svg}

\usepackage{graphicx}
\usepackage{amsmath}
\usepackage{scalerel}
\newcommand\reallywidehat[1]{\arraycolsep=0pt\relax%
\begin{array}{c}
\stretchto{
  \scaleto{
    \scalerel*[\widthof{\ensuremath{#1}}]{\kern-.5pt\bigwedge\kern-.5pt}
    {\rule[-\textheight/2]{1ex}{\textheight}} 
  }{\textheight} %
}{0.5ex}\\           
#1\\                 
\rule{-1ex}{0ex}
\end{array}
}

\title{Curiosity in exploring chemical space: Intrinsic rewards for deep molecular reinforcement learning}

\author{
  Luca A. Thiede \\
  Department of Physics, University of Göttingen, Germany\\
  Department of Computer Science, University of Toronto, Canada\\
  \texttt{luca.thiede@yahoo.com} \\
  \And
  Mario Krenn \\
  Chemical Physics Theory Group, Department of Chemistry, University of Toronto, Canada \\
  Department of Computer Science, University of Toronto, Canada\\
  Vector Institute for Artificial Intelligence, Toronto, Canada\\
  \texttt{mario.krenn@utoronto.ca} \\
  \AND
  AkshatKumar Nigam\\
  Department of Computer Science, University of Toronto, Canada\\
  Chemical Physics Theory Group, Department of Chemistry, University of Toronto, Canada\\
  \texttt{akshat.nigam@mail.utoronto.ca} \\
  \And
  Alán Aspuru-Guzik \\
  Chemical Physics Theory Group, Department of Chemistry, University of Toronto, Canada \\
  Department of Computer Science, University of Toronto, Canada\\
  Vector Institute for Artificial Intelligence, Toronto, Canada\\
  Canadian Institute for Advanced Research (CIFAR) Lebovic Fellow, Toronto, Canada\\
  \texttt{alan@aspuru.com} \\
}

\begin{document}

\maketitle

\begin{abstract}
Computer-aided design of molecules has the potential to disrupt the field of drug and material discovery. Machine learning, and deep learning, in particular, have been topics where the field has been developing at a rapid pace. Reinforcement learning is a particularly promising approach since it allows for molecular design without prior knowledge. However, the search space is vast and efficient exploration is desirable when using reinforcement learning agents. In this study, we propose an algorithm to aid efficient exploration. The algorithm is inspired by a concept known in the literature as {\sl curiosity}. We show on three benchmarks that a \textit{curious} agent finds better performing molecules. This indicates an exciting new research direction for reinforcement learning agents that can explore the chemical space out of their own motivation. This has the potential to eventually lead to unexpected new molecules that no human has thought about so far. 
\end{abstract}

\section{Introduction}\label{sec1}
The development of new drugs and functional materials is an important but expensive undertaking. To bring the cost down, the scientific community introduced computational screening methods that frame the discovery process as an optimization problem of desired properties over a large molecule database or chemical space. This is also known as the inverse molecular design problem \cite{sanchez2018inverse,gromski2019explore}. However, the search space is enormous \cite{search_space_enormous, search_space_enormous2} rendering exhaustive search of all possible molecules unfeasible. Thus, various artificial intelligence (A.I). approaches have been developed to tackle this problem, including variational autoencoders (VAEs) \cite{vae, vae_junction}, recurrent neural networks \cite{mol_rnn, mol_rnn2, mol_rnn3} and generative adversarial networks (GANs) \cite{organ, organic}. These approaches are very promising, but they require a sometimes large, training dataset. However, not for every class of molecules, such a training dataset exists, or it is very small. Furthermore, the use of a dataset biases the model and thus makes it unlikely to find interesting molecules outside of the given data distribution.

Other approaches like genetic algorithms \cite{genetic, jensen2019graph, deriver} or reinforcement learning (RL) allows designs potentially far away from any known data distribution \cite{rl1, rl2, gaudinexploring}. Instead of a dataset, only a reward function is needed, that measures how good a generated molecule is. However, due to the vast chemical space, efficient exploration is necessary.

Here we take inspiration from the field of RL for video games, where the use of a method called \textit{curiosity} \cite{curiosity_original} has led to impressive results without access to actual rewards from an environment \cite{burda2018large}. Curiosity falls under the wider category of intrinsic motivation techniques \cite{curiosity_original, intrinsic_motivation_survey, schmiddi}, which are loosely modeled after human curiosity. In this work,  we propose for the first time the use of intrinsic motivation for molecular design and show that the more curious agents in our study perform better than their less curious counterparts on three commonly used benchmarks.

\subsection{Reinforcement Learning basics}
Reinforcement Learning is a technique used to find a policy $\pi_\theta$ parameterized by $\theta$ that maximizes state-action trajectories in an environment. Formally, the environment is described as a Markov decision process $\mathcal{M} = (\mathcal{S}, \mathcal{A}, \mathcal{T}, \mu_0, \gamma, R, T)$. Here, $\mathcal{S}$ is the state space, $\mathcal{A}$ is the action space, $\mathcal{T} : \mathcal{S} \times \mathcal{A} \rightarrow \mathcal{S}$ is the transition function, $\mu_0$ is the initial state distribution, $\gamma \in (0,1]$ is the discount factor. $R : \mathcal{S} \times \mathcal{A} \rightarrow \mathbb{R}$ is the reward function. We define $r_t := R(s_t,a_t)$ and $T$ a the maximal length of an episode. For every policy $\pi$, the expected reward is defined as the reward that an agent will collect when it is in a certain state $V^{\pi}(s_{t^*}) = \mathbb{E}_\pi (\sum_{t=t^*}^T \gamma^t R(s_t, a_t|s_{t^*}))$. This quantity is also called the value of the state $s_{t^*}$. Analogously we define the Q value of an action in a state as $Q^{\pi}(s_{t^*}, a_{t^*}) = \mathbb{E}_\pi (\sum_{t=t^*}^T \gamma^t R(s_t, a_t| s_{t^*},a_{t^*}))$. The goal is to find a policy so that $J(\theta) = \mathbb{E}_{s_0 \sim \mu_0}(V^\pi(s_0))$ is maximized.\\

There are many different ways to train a policy. Throughout this paper, we will use a policy gradient method, in particular proximal policy optimization (PPO). We use the hyperparameter provided by the original paper \cite{ppo} which were tuned on atari games. The authors in \cite{benevolentai_study} conducted a thorough hyperparameter search and confirmed that these hyperparameters are also nearly optimal for molecular design and also other authors \cite{rl2} reported that they could not find better settings.

\section{Reinforcement Learning for molecular design}
For molecular design, we define the state $s_t$ as the SELFIES \cite{selfies} string that is so far constructed. SELFIES is a 100\% robust string-based representation of molecules. This is in contrast to SMILES or other string-based representations, which frequently produce semantically or syntactically invalid strings.  The robustness  of SELFIES is advantageous as our RL agent produces in every step a valid molecule. Thereby, the training doesn't require any post-processing or filters, thus is simpler. The action $a_t$ is the next character to be appended to the string. The molecule is finished either when the max number of steps is reached, which we set to 35 throughout our experiments, or the agents use the $\textit{[STOP]}$ symbol. \\
For some property $p$ that we wish to optimize, and by denoting the molecule at time step $t$ as $\text{mol}$, the reward at every time step can be formulated in two ways. Either as

\begin{align}
    r_t &= \begin{cases}
      p(\text{mol}(T)), & \text{SELFIES string finished} \\
      0, & \text{otherwise}  
    \end{cases} \label{eq:reward1}
\end{align}
or as 
\begin{align}
    r_t &= p(\text{mol}(t)) - p(\text{mol}(t-1)) \label{eq:reward2}
\end{align}
For both formulations the cumulative reward is the property of the final molecule ($\sum^T_{t=0} \gamma^t r_t  = p(\text{mol}(t))$) for $\gamma=1$. The first formulation is sparse, but only needs one evaluation of the reward per trajectory while the second formulation is more dense and therefore more informative, but also requires a lot more calculations of the reward function. \\
The chemical space is huge, and therefore efficient exploration is necessary to find good solutions. In the next section we discuss a technique in reinforcement learning literature that is known to help with better exploration and adapt it to our use case.

\begin{figure} 
  \centering
  \includegraphics[width=\textwidth,height=\textheight,keepaspectratio]{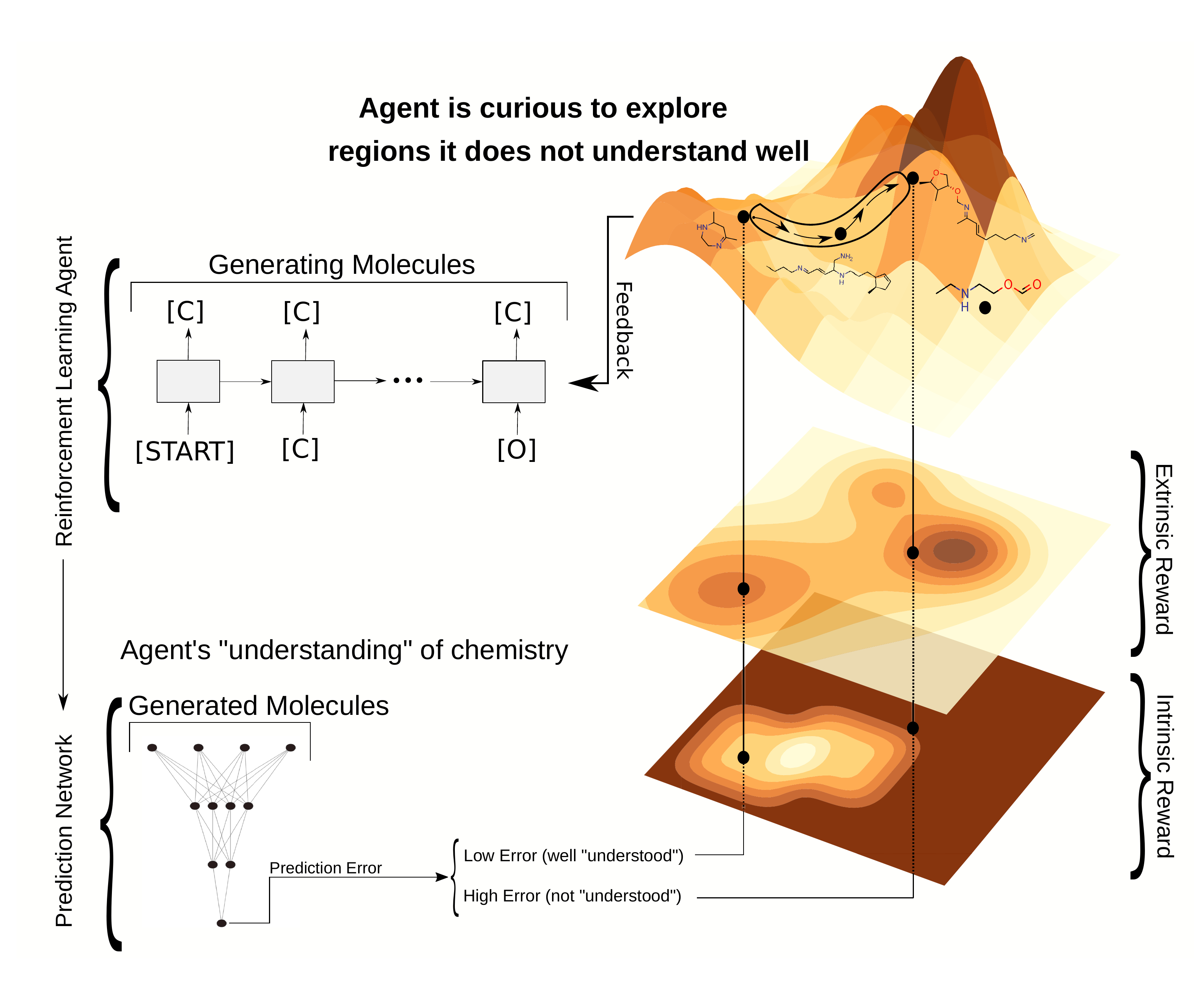}
  \caption{An illustration of curiosity: The agent generates molecules and gives them to the property prediction network. Initially, the predictions are wrong everywhere in the chemical space, but over time the network learns to predict the properties of molecules it has already encountered well. By using the prediction error as an intrinsic reward, the agent is incentivised to move to regions, it has not yet explored, because the prediction network will make more errors there. The combination of the intrinsic and extrinsic rewards is then used as feedback to train the agent.}
  \label{fig:curiosity_memory_explanation}
\end{figure}

\section{Related Work}
The literature on reinforcement learning often distinguishes between intrinsic and extrinsic rewards. An extrinsic reward is anything that comes directly from the environment. Intrinsic rewards are any rewards that are generated by the agent itself. Pathak et. al \cite{curiosity_original} introduce an intrinsic reward called curiosity. Curiosity guides the exploration of an agent into regions of the state space, where it has not understood the effect of its actions on the environment. They introduced a separate neural network that tries to predict the next state the agent will be in given the action it took. Then, the total reward the agent gets is
\begin{align}
    r_\text{total}(t) = r_\text{extrinsic}(t) + \alpha r_\text{intrinsic}(t) \label{eq:intrinsic_loss}
\end{align}
where  $r_\text{extrinsic}(t)$ is the normal reward provided by the environment, and $r_\text{intrinsic}(t)$ is the error of the prediction from the newly introduced neural network. The prediction network can also be seen as implicitly storing information in its weights about which areas of the state-action space have already been visited, since the more often the agent is in a certain region of that space, the smaller the prediction error is going to be. By exploiting this information, the agent will continue exploring new regions and not get stuck in local optima. This situation is depicted in Figure \ref{fig:curiosity_memory_explanation}.

\section{Curiosity for molecular design}
Our goal is to adapt the work described in the previous section to molecular design. For this, we use a prediction network that predicts the property of the next molecule and use the prediction error as the intrinsic reward (see Figure \ref{fig:curiosity_memory_explanation}). In order to test different variations of this idea, we formulate it generally as:
\begin{align}
     r_\text{intrinsic}(t) = \text{dist}\left(\reallywidehat{\text{p}}\left(\text{mol}\left(t, \theta\right), \eta\right), \text{p}\left(\text{mol}\left(t, \theta\right)\right)\right) \label{eq:general_form}
\end{align}
Here $\hat{\text{p}}(\cdot, \eta)$ is the prediction network parameterized by $\eta$ that tries to predict the real value of the considered property $p$ of the molecule. $\text{mol}(t, \theta)$ is the molecule the agent parameterized by $\theta$ generates at time step $t$ and $\text{dist}(\cdot, \cdot)$ is a distance metric, for example $L1$ or $L2$. 
We also try an alternative formulation that we call "greedy curiosity". It uses a mask function to be curious only in promising direction (see appendix \ref{appendix:greedy_curiosity}).
For training the predictor network we consider two options: Either we update the predictor network each time the agent generates new molecules, or we collect them in a buffer and train on the whole buffer (see appendix \ref{appendix:training_details}).

Unlike Pathak et. al. \cite{curiosity_original}, we do not predict the next state. The reason is, that given the current state (the string so far), and the next action (the character to append), predicting the next state (the string so far with the new character appended) does not require to learn anything about the chemical space. Instead of storing information in the state-action space, our prediction network only remembers regions of the state space the agent visited. \\
We also consider a very simple alternative where we explicitly store the last $N$ molecules into a buffer and calculate the average Tanimoto Similarity (TS) of the Morgan Fingerprints (MF) \cite{fingerprints}: 
\begin{align}
    r_\text{intrinsic, alternative}(t) = - \frac{1}{N}\sum_{i=0}^N\text{TS}(\text{MF}(\text{mol}(t)), \text{MF}(\text{mol}_i))
\end{align}
This approach explicitly makes a choice of a particular distance metric. In contrast, curiosity defines one implicitly, because molecules with a low prediction error are close to previously encountered molecules in the learned problem-specific feature space of the predictor network. Since this feature space is by construction useful for predicting the property that the agent ought to optimize, we hope that the induced similarity metric is more useful for the optimization task at hand.

\begin{figure} 
  \centering
  \includegraphics[width=\textwidth,height=\textheight,keepaspectratio]{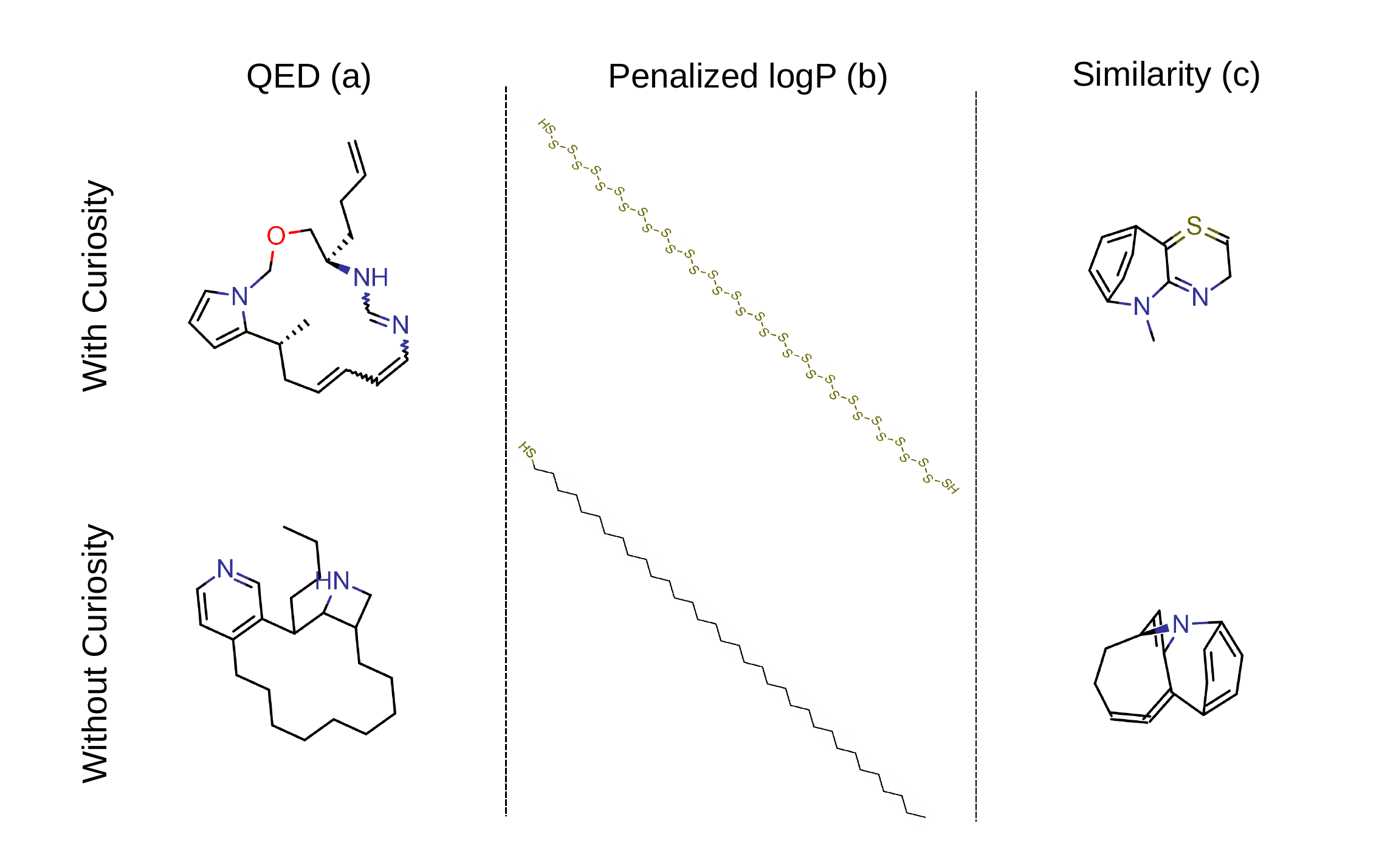}
  \caption{Best molecule generated on the different tasks. Tasks from top to bottom: QED, penalized logP, Similarity. Left: Best molecule generated by the curious agent. Right: Best molecule generated by a non curious agent. Note that no chemical stability or synthesizabitly filters were employed on this work, so the structures may look a bit strange to organic chemists.}
  \label{fig:molecules}
\end{figure}

\section{Experiments}
We test our method on 3 different tasks. The three tasks are optimizing for Quantitative Estimate of Druglikeness \cite{qed_paper} (QED), penalized logP \cite{vae} (plogP), and similarity (in terms of Tanimoto similarity of Morgan fingerprints) to the target molecule Celecoxib, a task from the Guacamol benchmark \cite{guacamol}.  
QED is a statistical measure to quantify if a specific set of 8 properties of a molecule fit those of 771 approved orally administered drugs. Penalized logP rewards the logP value and synthetic accessibility score of a molecule, and penalizes its ring count. This property is often used in drug design literature. For penalized logP the best-known optimum is the sulfur chain \cite{genetic}. It turned out that the carbon chain was a good local minimum all agents were getting stuck in. Therefore we were providing the [S] symbol as the initial state so that some of the agents were able to find the sulfur chain.\\
For each task we train 3 agents for all possible combinations of $\alpha = \{0, 0.01, 0.1, 1\}$, $r_\text{intrinsic, alternative}$, the distance metric $L1/L2$ and whether or not Greedy curiosity and the buffer are used. \\
We divide the extrinsic and intrinsic rewards by their respective running averages so that their influence is on the same order of magnitude and the scaling factor $\alpha$ is comparable across experiments. The agent consists of an LSTM \cite{LSTM} with 64 units to encode the state and a linear network to predict the action. The prediction network has the same architecture, but does not share any neural network weights. \\

\subsection{Results}
The averaged results of the 3 runs for the best performing hyperparameter sets are shown in Table \ref{tab:results_qed} - \ref{tab:results_similarity} for the 3 different tasks. Additionally the best value for an agent without curiosity ($\alpha = 0$) and the best value for an agent using $r_\text{intrinsic, alternative}$ are shown.

\begin{table}[h!]
 \begin{subtable}[h]{0.45\textwidth}
 \centering
\begin{tabular}{|l|l|l|l|l|l|l|l|}
\hline
 Curiosity weight $\alpha$ & $r_\text{intrinsic, alternative}$ & dist  & Greedy curiosity  & Use buffer & Best QED \\ \hline \hline
\textbf{1}        & \textbf{False}                                    & $\boldsymbol{L_2}$ & \textbf{False}                     & \textbf{False}      & \textbf{0.918}   \\ \hline
1        & False                                   & $L_1$ & False                    & False      & 0.916    \\ \hline
0.1      & False                                    & $L_2$ & False                    & False      & 0.898    \\ \hline
0        & -                                       & -     & -                         & -          & 0.889    \\ \hline
0.1      & True                                      & False & False                    & -          & 0.883    \\ \hline
\end{tabular}
\caption{Results for the QED task}
\label{tab:results_qed}
\end{subtable}
\hfill
    
\begin{subtable}[]{0.45\textwidth}
 \centering
\begin{tabular}{|l|l|l|l|l|l|l|l|}
\hline
Curiosity weight $\alpha$ & $r_\text{intrinsic, alternative}$  & dist  & Greedy curiosity & Use buffer & Best p. logP \\ \hline \hline
\textbf{1}        & \textbf{False}                                     & $\boldsymbol{L_2}$ & \textbf{False}                     & \textbf{False}      & \textbf{10.364}       \\ \hline
1        & False                                     & $L_1$ & False                     & False      & 10.364       \\ \hline
0        & -                                         & -     & -                         & -          & 9.580      \\ \hline
0.1      & True                                      & -     & -                         & -          & 9.580      \\ \hline
\end{tabular}
\caption{Results for the penalized logP task}
\label{tab:results_plogp}
\end{subtable}
\hfill

\begin{subtable}[]{0.45\textwidth}
\centering
\begin{tabular}{|l|l|l|l|l|l|l|l|}
\hline
Curiosity weight $\alpha$ & $r_\text{intrinsic, alternative}$ & dist  & Greedy curiosity & Use buffer & Best similarity \\ \hline \hline
\textbf{1}        & \textbf{False}                                    & $\boldsymbol{L_1}$ & \textbf{False}                     & \textbf{False}      & \textbf{0.239}          \\ \hline
1        & False                                    & $L_1$ & True                      & False      & 0.237          \\ \hline
1        & False                                    & $L_2$ & False                     & False      & 0.236          \\ \hline
0.1      & True                                     & $L_2$ & True                      & -          & 0.224          \\ \hline
0        & -                                        & -     & -                         & -          & 0.186          \\ \hline
\end{tabular}
\caption{Results for the similarity task}
\label{tab:results_similarity}
\end{subtable}
\caption{The best values of the generated molecules, averaged over the 3 runs for the 3 best performing hyperparameter settings over all tasks. Additionally the average best value of an agent without curiosity ($\alpha = 0$), and one that uses $r_\text{intrinsic, alternative}$ are shown. The best agents all used the intrinsic reward ($\alpha = 1$).}
\end{table}

The agents with curiosity perform the best, moreover, the best-performing agents always have the highest curiosity weight ($\alpha$) from all tested hyperparameter sets. For the pLogP task, only two agents, both of which use curiosity, have found the sulfur chain. This indicates that curiosity indeed can help to escape local optima. The alternative formulation of the intrinsic reward seems to help for the similarity task but not on the QED task and does not help to find the sulfur chain. The version where we optimize the predictor network after every step of the agent consistently performed better than the version where we used a buffer, which is probably due to the fact, that we trained the predictor only two times during the agent's lifetime. In Figure \ref{fig:molecules} the best generated molecule for each task are shown.

Please note that for this work we did not employ any synthesizability or stability screening of the final molecules. This results in structures that are not so pleasant to the eye of the trained chemist. Further work can be done to add these as suitable additional terms in the reinforcement learning framework or alteratively as post-selection filters.

\section{Conclusion}
In this work, we develop the first \textit{curious} agents in the domain of molecular design and show that they outperform their lesser curious competitors in three distinct molecular design tasks. Our results point towards a new, efficient RL-based exploration strategy for identifying high-performance molecules and compounds. Since the agent explores the chemical space without any prior knowledge, in the future this can lead to the discovery of new molecules completely different from the ones scientists know of today. Interesting future experiments could further investigate the curiosity rewards, and explore how they are related or can be combined with other exploration strategies, such as count based exploration \cite{countbased, countbased_2}, parameter space noise injection \cite{noise_injection}, Go-Explore \cite{goexplore} and others.

These positive results indicate that curiosity, and intrinsic rewards in general, have the potential to significantly improve de-novo molecular design. We imagine three very interesting domains of application. First, the exploration of the large chemical space could be improved leading to better solutions. Second, if scientists are interested in exceptionally unique solutions that are very different from known structures, curiosity-based optimization techniques might be a way to obtain these. Lastly, our main motivation is to explore the potential advantage of intrinsic curiosity-based rewards in RL environment. We show in three examples how curious agents outperform their non-curious friends. As future research, it will be interesting to see how these insights can be combined to build algorithms \cite{nigam_stoned} that compete in a wide variety of different molecular design tasks.

Importantly, to understand how to exploit the advantage of curiosity in realistic applications, we will need to train on other more complex molecular properties, as well as to make the model aware of synthesizability and stability constraints by adding filters \cite{butt2000toxicity} and/or numerical approximations \cite{Gao2020} of the feasible regions within the chemical space. These two aspects, in general, should be a major area of focus for most of the work in this field.

\section{Acknowledgements}
The authors thank Theophile Gaudin and Dr. Si Yue Guo for interesting discussions. A. A.-G. acknowledges generous support from the Canada 150 Research Chairs Program, Tata Steel, Anders G Froseth, and the Office of Naval Research. L.A.T. acknowledges support from Mitacs via project FR44234. M.K. acknowledges support from the Austrian Science Fund (FWF) through the Erwin Schr\"odinger fellowship No. J4309. We acknowledge supercomputing support from SciNet.

\FloatBarrier

\bibliographystyle{unsrt}
\bibliography{refs}

\begin{thebibliography}{}

\bibitem{sanchez2018inverse}
B. Sanchez-Lengeling and A. Aspuru-Guzik, Inverse molecular design using
  machine learning: Generative models for matter engineering. \textit{Science}
  \textbf{361}, 360--365 (2018).

\bibitem{gromski2019explore}
P.S. Gromski, A.B. Henson, J.M. Granda and L. Cronin, How to explore chemical
  space using algorithms and automation. \textit{Nature Reviews Chemistry}
  \textbf{3}, 119--128 (2019).

\bibitem{search_space_enormous}
K.L. Drew, H. Baiman, P. Khwaounjoo, B. Yu and J. Reynisson, {{S}ize estimation
  of chemical space: how big is it?}. \textit{J Pharm Pharmacol} \textbf{64},
  490--495 (2012).

\bibitem{search_space_enormous2}
P.G. Polishchuk, T.I. Madzhidov and A. Varnek, Estimation of the size of
  drug-like chemical space based on GDB-17 data. \textit{Journal of
  computer-aided molecular design} \textbf{27}, 675--679 (2013).

\bibitem{vae}
R. G{\'o}mez-Bombarelli, J.N. Wei, D. Duvenaud, J.M. Hern{\'a}ndez-Lobato, B.
  S{\'a}nchez-Lengeling, D. Sheberla, J. Aguilera-Iparraguirre, T.D. Hirzel,
  R.P. Adams and A. Aspuru-Guzik, Automatic chemical design using a data-driven
  continuous representation of molecules. \textit{ACS central science}
  \textbf{4}, 268--276 (2018).

\bibitem{vae_junction}
W. Jin, R. Barzilay and T. Jaakkola, Junction Tree Variational Autoencoder for
  Molecular Graph Generation. \textit{International Conference on Machine
  Learning} 2323--2332 (2018).

\bibitem{mol_rnn}
E.J. Bjerrum and R. Threlfall, Molecular generation with recurrent neural
  networks (RNNs). \textit{arXiv:1705.04612} (2017).

\bibitem{mol_rnn2}
M.H.S. Segler, T. Kogej, C. Tyrchan and M.P. Waller, Generating Focused
  Molecule Libraries for Drug Discovery with Recurrent Neural Networks.
  \textit{ACS Central Science} \textbf{4}, 120-131 (2018).

\bibitem{mol_rnn3}
P. Ertl, R. Lewis, E. Martin and V. Polyakov, In silico generation of novel,
  drug-like chemical matter using the LSTM neural network.
  \textit{arXiv:1712.07449} (2017).

\bibitem{organ}
G.L. Guimaraes, B. Sanchez-Lengeling, C. Outeiral, P.L.C. Farias and A.
  Aspuru-Guzik, Objective-reinforced generative adversarial networks (ORGAN)
  for sequence generation models. \textit{arXiv:1705.10843} (2017).

\bibitem{organic}
B. Sanchez-Lengeling, C. Outeiral, G.L. Guimaraes and A. Aspuru-Guzik,
  Optimizing distributions over molecular space. An objective-reinforced
  generative adversarial network for inverse-design chemistry (ORGANIC).
  \textit{chemrXiv:5309668} .

\bibitem{genetic}
A. Nigam, P. Friederich, M. Krenn and A. Aspuru-Guzik, Augmenting genetic
  algorithms with deep neural networks for exploring the chemical space.
  \textit{arXiv:1909.11655} (2019).

\bibitem{jensen2019graph}
J.H. Jensen, A graph-based genetic algorithm and generative model/Monte Carlo
  tree search for the exploration of chemical space. \textit{Chemical science}
  \textbf{10}, 3567--3572 (2019).

\bibitem{deriver}
S. Reeves, B. DiFrancesco, V. Shahani, S. MacKinnon, A. Windemuth and A.E.
  Brereton, Assessing Methods and Obstacles in Chemical Space Exploration.
  \textit{Applied AI Letters, ail2.17} (2020).

\bibitem{rl1}
M. Olivecrona, T. Blaschke, O. Engkvist and H. Chen, Molecular de-novo design
  through deep reinforcement learning. \textit{Journal of cheminformatics}
  \textbf{9}, 48 (2017).

\bibitem{rl2}
J. You, B. Liu, Z. Ying, V. Pande and J. Leskovec, Graph convolutional policy
  network for goal-directed molecular graph generation. \textit{Advances in
  neural information processing systems} 6410--6421 (2018).

\bibitem{gaudinexploring}
T. Gaudin, A. Nigam and A. Aspuru-Guzik, Exploring the chemical space without
  bias: data-free molecule generation with DQN and SELFIES. \textit{Second
  Workshop on Machine Learning and the Physical Sciences NeurIPS} .

\bibitem{curiosity_original}
D. Pathak, P. Agrawal, A.A. Efros and T. Darrell, Curiosity-driven exploration
  by self-supervised prediction. \textit{Proceedings of the IEEE Conference on
  Computer Vision and Pattern Recognition Workshops} 16--17 (2017).

\bibitem{burda2018large}
Y. Burda, H. Edwards, D. Pathak, A. Storkey, T. Darrell and A.A. Efros,
  Large-scale study of curiosity-driven learning. \textit{arXiv:1808.04355}
  (2018).

\bibitem{intrinsic_motivation_survey}
A. Aubret, L. Matignon and S. Hassas, A survey on intrinsic motivation in
  reinforcement learning. \textit{arXiv:1908.06976} (2019).

\bibitem{schmiddi}
J. Schmidhuber, Formal theory of creativity, fun, and intrinsic motivation
  (1990--2010). \textit{IEEE Transactions on Autonomous Mental Development}
  \textbf{2}, 230--247 (2010).

\bibitem{ppo}
J. Schulman, F. Wolski, P. Dhariwal, A. Radford and O. Klimov, Proximal policy
  optimization algorithms. \textit{arXiv:1707.06347} (2017).

\bibitem{benevolentai_study}
D. Neil, M.H.S. Segler, L. Guasch, M. Ahmed, D. Plumbley, M. Sellwood and N.
  Brown, Exploring Deep Recurrent Models with Reinforcement Learning for
  Molecule Design. \textit{ICLR} (2018).

\bibitem{selfies}
M. Krenn, F. Hase, A. Nigam, P. Friederich and A. Aspuru-Guzik,
  Self-Referencing Embedded Strings (SELFIES): A 100\% robust molecular string
  representation. \textit{Machine Learning: Science and Technology} \textbf{1},
  045024 (2020).

\bibitem{fingerprints}
D. Rogers and M. Hahn, Extended-Connectivity Fingerprints. \textit{Journal of
  Chemical Information and Modeling} \textbf{50}, 742-754 (2010).

\bibitem{qed_paper}
G.R. Bickerton, G.V. Paolini, J. Besnard, S. Muresan and A.L. Hopkins,
  Quantifying the chemical beauty of drugs. \textit{Nature chemistry}
  \textbf{4}, 90-98 (2012).

\bibitem{guacamol}
N. Brown, M. Fiscato, M.H. Segler and A.C. Vaucher, GuacaMol: benchmarking
  models for de novo molecular design. \textit{Journal of chemical information
  and modeling} \textbf{59}, 1096--1108 (2019).

\bibitem{LSTM}
S. Hochreiter and J. Schmidhuber, Long Short-Term Memory. \textit{Neural
  Comput.} \textbf{9}, 1735–1780 (1997).

\bibitem{countbased}
M. Bellemare, S. Srinivasan, G. Ostrovski, T. Schaul, D. Saxton and R. Munos,
  Unifying count-based exploration and intrinsic motivation. \textit{Advances
  in neural information processing systems} \textbf{29}, 1471--1479 (2016).

\bibitem{countbased_2}
H. Tang, R. Houthooft, D. Foote, A. Stooke, O.X. Chen, Y. Duan, J. Schulman, F.
  DeTurck and P. Abbeel, \# exploration: A study of count-based exploration for
  deep reinforcement learning. \textit{Advances in neural information
  processing systems} 2753--2762 (2017).

\bibitem{noise_injection}
M. Plappert, R. Houthooft, P. Dhariwal, S. Sidor, R.Y. Chen, X. Chen, T.
  Asfour, P. Abbeel and M. Andrychowicz, Parameter space noise for exploration.
  \textit{arXiv:1706.01905} (2017).

\bibitem{goexplore}
A. Ecoffet, J. Huizinga, J. Lehman, K.O. Stanley and J. Clune, Go-explore: a
  new approach for hard-exploration problems. \textit{arXiv:1901.10995} (2019).

\bibitem{nigam_stoned}
A. Nigam, R. Pollice, M. Krenn, G. Passos~Gomes and A. Aspuru-Guzik, Beyond
  Generative Models: Superfast Traversal, Optimization, Novelty, Exploration
  and Discovery (STONED) Algorithm for Molecules using SELFIES.
  \textit{ChemRxiv:13383266} (2020).

\bibitem{butt2000toxicity}
S.T. Butt and T. Christensen, Toxicity and phototoxicity of chemical sun
  filters. \textit{Radiation protection dosimetry} \textbf{91}, 283--286
  (2000).

\bibitem{Gao2020}
W. Gao and C.W. Coley, The Synthesizability of Molecules Proposed by Generative
  Models. \textit{Journal of Chemical Information and Modeling} (2020).

\end{thebibliography}

\appendix
\section{Greedy curiosity}
\label{appendix:greedy_curiosity}
We tried out another alternative formulation that we call greedy curiosity. The intuition behind greedy curiosity is that the agent is only curious in promising directions and ignores directions in the state space which have a high prediction error due to a very low reward. We achieve this by multiplying the prediction error by a mask function. Equation \eqref{eq:general_form} becomes
\begin{align}
     r_\text{intrinsic}(t) = \text{dist}\left(\reallywidehat{\text{p}}\left(\text{mol}\left(t, \theta\right), \eta\right), \text{p}\left(\text{mol}\left(t, \theta\right)\right)\right) \cdot \text{mask}\left(\text{mol}\left(t, \theta\right)_{1...\text{batch size}}\right)
\end{align}
The mask function $\text{mask}(\text{mol}(t, \theta)_{1...\text{batch size}})$  gets the whole batch of molecules as input, and masks off the curiosity reward for all molecules which target properties are worse than the average in the batch:
\begin{align}
    \text{mask}(\text{mol}(t, \theta)_{i=1...\text{batch size}})_i = \begin{cases}
      1, & \text{if}\ \text{p}(\text{mol}(t, \theta)_i) - \text{mean}({\text{p}(\text{mol}(t, \theta)_{1...\text{batch size}}))} > 0 \\
      0, & \text{otherwise}
    \end{cases}
\end{align}

\section{Curiosity training details}
\label{appendix:training_details}
For training the predictor network we consider two options: The first is to update the predictor network after every episode with the new batch of generated samples. A potential downside of this is, that the predictor might forget about older samples. The second option is to use a buffer and collect and train the predictor on all samples. One can either reinitialize the predictor every time before training, which makes it very resource-intensive, or once can do warm starts. However, old samples will be seen more often than new samples, leading to overfitting. Thus we opted for reinitializing and training the predictor only two times, after 200, and again after 500 episodes. In our setup, when using  equation \eqref{eq:reward1} we have access to the property of the molecule at every time step. Thus, we can use all molecules at every time step to train the prediction network. If we use equation \eqref{eq:reward2} instead, we have access to molecules only at the end of each trajectory, and therefore use only the last molecule of the trajectory to train the predictor. \\

The alternative reward formulation has the downside, that the runtime scales with the size of the memory and becomes the most time consuming step even for a small number of stored molecules $N$. Therefore we were only storing the last 2 batches of molecules in the memory which was giving it about the same computational budget as the prediction network. 
\end{document}